\documentclass{bmvc2k}

\newlength\savedwidth
\newcommand{\whline}{\noalign{\global\savedwidth\arrayrulewidth
                            \global\arrayrulewidth 1.5pt}%
                   \hline
                   \noalign{\global\arrayrulewidth\savedwidth}}

\expandafter\def\expandafter\normalsize\expandafter{%
    \normalsize
   \setlength\abovedisplayskip{5pt}
  \setlength\belowdisplayskip{5pt}
  \setlength\abovedisplayshortskip{5pt}
   \setlength\belowdisplayshortskip{5pt}
}

\usepackage[floatrowsep=none]{floatrow}
\newfloatcommand{capbtabbox}{table}[][\FBwidth]
\usepackage{bm}
\usepackage{enumitem}
\usepackage{floatrow}
\usepackage{multirow}
\usepackage{subfigure}
\usepackage[ruled]{algorithm2e}
\usepackage{graphicx}
\usepackage{amsmath,amssymb} 
\usepackage{tabularx}
\usepackage{colortbl}
\definecolor{Gray}{gray}{0.9}

\title{Deep Reinforcement Learning Attention Selection for Person Re-Identification}

\addauthor{Xu Lan}{x.lan@qmul.ac.uk}{1}
\addauthor{Hanxiao Wang}{hanxiao.wang@qmul.ac.uk}{1}
\addauthor{Shaogang Gong}{s.gong@qmul.ac.uk}{2}
\addauthor{Xiatian Zhu}{xiatian.zhu@qmul.ac.uk}{2}
\addinstitution{
 Vision Group\\
 School of Electronic Engineering and
 Computer Science\\
Queen Mary University of London\\
 London  E1 4NS UK
}

\runninghead{Student, Prof, Collaborator}{BMVC Author Guidelines}

\begin{document}
\maketitle
\begin{abstract}
Existing person re-identification (re-id) methods assume
the provision of accurately cropped person bounding boxes with minimum
background noise, mostly by {\em manually cropping}. This is
significantly breached in practice when person bounding boxes must be
{\em detected automatically} given a very large number of images and/or
videos processed.
Compared to carefully cropped manually, auto-detected bounding boxes
are far less accurate with random amount of background clutter
which can degrade notably person re-id matching accuracy. 
In this work, we develop a joint learning deep model that optimises
person re-id attention selection within any auto-detected person
bounding boxes by {\em reinforcement learning} of background clutter minimisation
subject to re-id label pairwise constraints. 
Specifically, we formulate a novel unified re-id architecture called 
{\bf I}dentity {\bf D}iscriminativ{\bf E} {\bf A}ttention reinforcement {\bf L}earning (IDEAL)
to accurately select re-id attention in auto-detected bounding boxes
for optimising re-id performance. Our model can improve re-id accuracy
comparable to that from exhaustive human manual cropping of
bounding boxes with additional advantages from identity discriminative
attention selection that specially benefits re-id tasks beyond human knowledge.
Extensive comparative evaluations demonstrate the re-id advantages of
the proposed IDEAL model over a wide range of state-of-the-art re-id
methods on two auto-detected re-id benchmarks CUHK03 and Market-1501. 

\end{abstract}


\section{Introduction}
\label{sec:intro}

\begin{figure}
\centering
\includegraphics[width=1\linewidth]{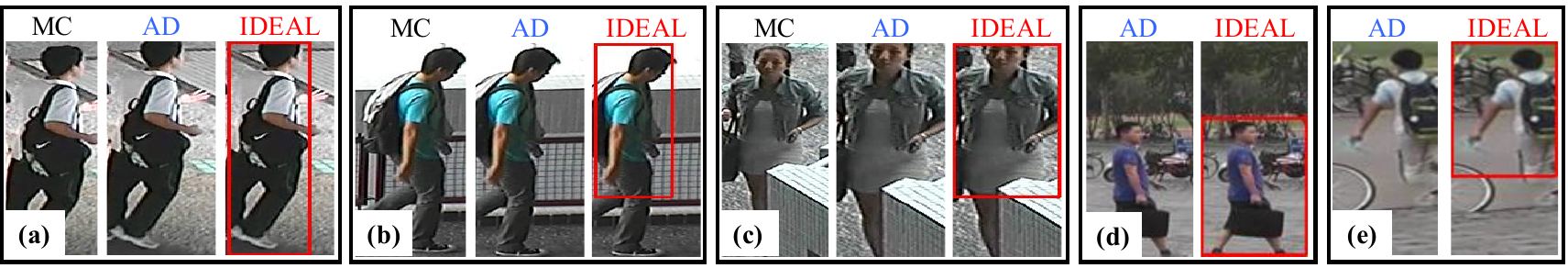}
\vskip -0.45cm
\caption{\footnotesize
	Comparisons of person bounding boxes by manually cropping (MC), automatically detecting (AD),
	and identity discriminative attention reinforcement learning (IDEAL). 
	Often AD contains more background
        clutter (a,d,e). Both AD and MC may suffer from occlusion (c),
        or a lack of identity discriminative attention selection (b).
}
\label{Fig:1}
\end{figure}

Person re-identification (re-id) aims at searching people across non-overlapping camera
views distributed at different locations by matching person bounding box images~\cite{gong2014person}. 
In real-world re-id scenarios, 
{\em automatic person detection} \cite{felzenszwalb2010object} is {\em
  essential} for re-id to scale up to large size data, e.g. more
recent re-id benchmarks CUHK03 \cite{Li_DeepReID_2014b} and 
Market-1501 \cite{zheng2015scalable}. Most existing re-id test
datasets~(Table \ref{tab:datasets}) are {\em manually cropped}, as in VIPeR \cite{VIPeR} and
iLIDS \cite{RankSVMReId_BMVC10}, thus they do not fully address the re-id
challenge in practice. 
However, 
auto-detected bounding boxes are not optimised for re-id tasks due to 
potentially more background clutter, occlusion, missing body part, and
inaccurate bounding box alignment (Fig. \ref{Fig:1}).
This is evident from that the rank-1 re-id rate on CUHK03 drops
significantly from 61.6\%\ on manually-cropped to
53.4\% on auto-detected bounding boxes by state-of-the-art
hand-crafted models~\cite{wang2016highly}, that is, a 8.2\% rank-1 drop;
and from 75.3\% on manually-cropped \cite{xiao2016learning} to 68.1\%
on auto-detected \cite{Gated_SCNN} by state-of-the-art deep learning
models, that is, a 7.2\% rank-1 drop.
Moreover, currently reported ``auto-detected'' re-id performances on both
CUHK03 and Market-1501 have further benefited from 
artifical {\em human-in-the-loop cleaning process}, which discarded ``bad''
detections with $<50\%$ IOU (intersection over union) overlap with
corresponding manually cropped bounding boxes. Poorer detection
bounding boxes are considered as ``distractors'' in Market-1501 and
not given re-id labelled data for model learning.
In this context, there is a need for {\em attention selection} within
auto-detected bounding boxes as an integral part of learning to
optimise person re-id accuracy in a fully automated process.

\vskip -0.3cm
\begin{table}[h!]
	\centering
	\footnotesize
	\renewcommand{\arraystretch}{1}
	\setlength{\tabcolsep}{0.08cm}
	\caption{\footnotesize
	Person re-id datasets with/without auto-detection. MC: Manual Cropping; AD: Automatic Detection.
	}
	\vskip-0.3cm
	\label{tab:datasets}
	\scalebox{1}{
		\begin{tabular}{|c|cccccc|}
			\whline
			Dataset  & VIPeR \cite{VIPeR} 
			& GRID \cite{GRID} 
			& iLIDS \cite{RankSVMReId_BMVC10}
			& CAVIAR4ReID \cite{cheng2011custom}
			& CUHK03 \cite{Li_DeepReID_2014b} & Market-1501 \cite{zheng2015scalable} 
			\\ \hline \hline 
			Year
			& 2007 & 2009 & 2010 & 2011 & 2014 & 2015 
			\\ \hline
			Annotation
			& MC & MC & MC & MC
			& MC+AD & AD 
			\\ \hline
			Identities
			& 632 & 250 & 119 & 72 & 1,360 & 1,501
			\\ 
			Images 
			& 1,264 & 1,275 & 476 & 1,221 & 28,192 & 32,668
			\\ 
			\whline
	\end{tabular}}
\vspace{-0.3cm}
\end{table}
%
%

\begin{figure}[th!]
	\centering
	\includegraphics[width=0.99\linewidth]{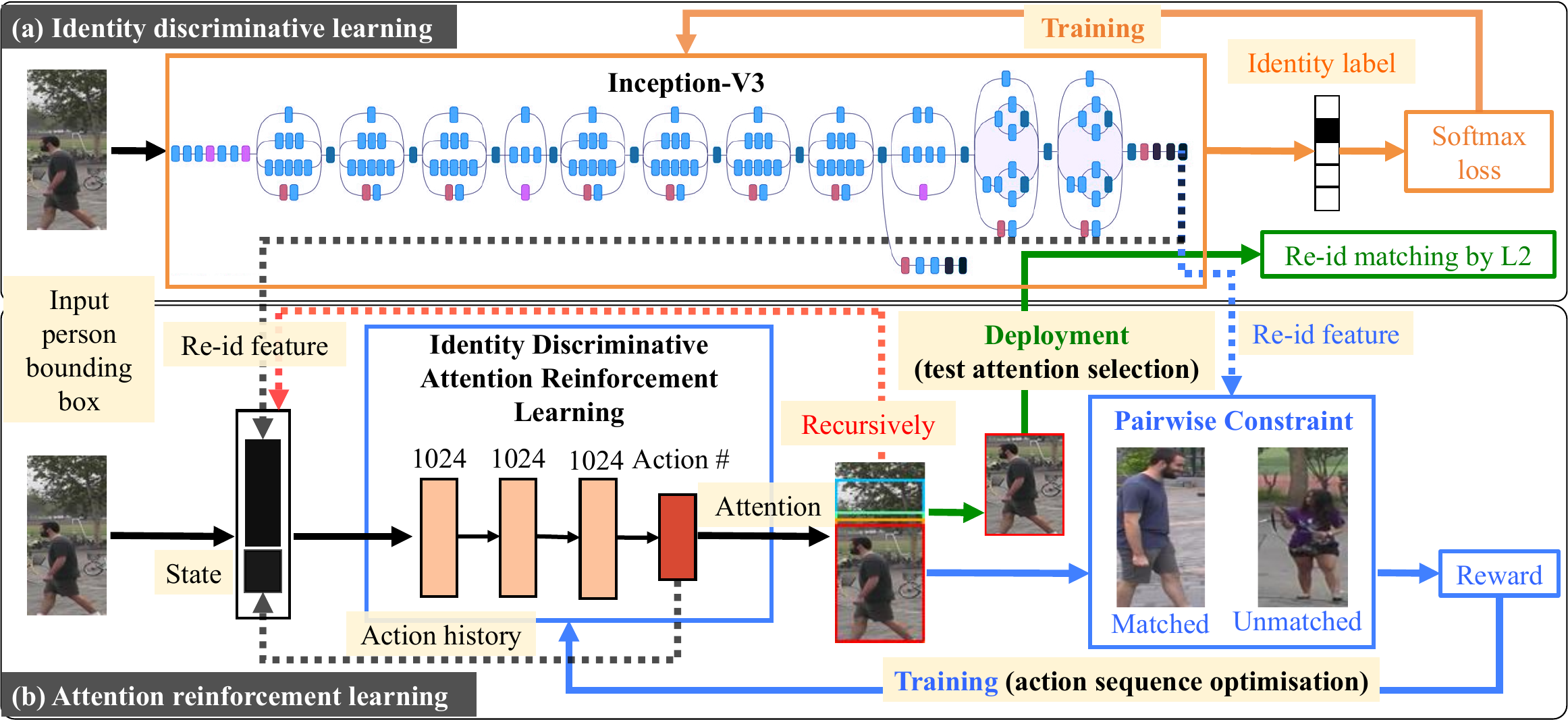}
	\vskip -0.3cm
	\caption{\footnotesize The IDEAL reinforcement learning attention selection model. 
	(a) An identity discriminative learning branch based on the deep
          Inception-V3 network optimised by a multi-classification
          softmax loss (orange arrows). 
	(b) An attention reinforcement learning branch designed as a deep Q-network
	optimised by re-id class label constraints in the deep feature
        space from branch (a) (blue arrows). 
	For model deployment, the trained attention branch
        (b) computes the optimal attention regions for
        each probe and all the gallery images, extract the deep
        features from these optimal attention regions in the multi-class
        re-id branch (a) and perform L2 distance matching (green arrows).
}
	\label{Fig:architecture}
\end{figure}

There is very little attempt in the literature for solving
this problem of attention selection within auto-detected bounding
boxes for optimising person re-id, except a related recent study on joint
learning of person detection and re-id \cite{xiao2016end}. Our
approach however differs from that by operating on any third party detectors
{\em independently} so to benefit continuously from a wide range of
detectors being rapidly developed by the wider community. 
Other related possible strategies include 
local patch calibration for mitigating misalignment in pairwise image matching~\cite{SalienceReId_CVPR13,shen2015person,zheng2015partial,Gated_SCNN}
and local saliency learning for region soft-selective matching~\cite{zhao2013person,hanxiao2014GTS,SalienceReId_CVPR13,liu2016end}.
These methods have shown to reduce the effects from viewpoint and
human pose change on re-id accuracy. However, {\em all} of them assume that person bounding boxes
are reasonably accurate. 
In this work, we consider the problem of optimising attention selection
within any auto-detected person bounding boxes for maximising re-id tasks.
The {\bf contributions} of this study are:
{\bf (1)} We formulate
a novel {\bf I}dentity {\bf D}iscriminativ{\bf E} {\bf A}ttention reinforcement {\bf L}earning 
(IDEAL) model for attention selection {\em post-detection} given re-id discriminative constraints.
Specifically, IDEAL is designed to 
locate automatically identity-sensitive attention regions 
within auto-detected bounding boxes by
optimising recursively attending actions using {\em reinforcement
learning} subject to a reward function on satisfying re-id pairwise label
constraints 
(Fig. \ref{Fig:architecture}).
In contrast to existing saliency 
selection methods, this global attention selection approach is more scalable 
in practice. This is because that most saliency models are local-patch
based and assume good inter-image alignment, or it requires complex
manipulation of local patch correspondence {\em independently}, difficult to scale. The IDEAL
attention model is directly estimated under a discriminative re-id matching criterion
to jointly maximise a reinforcement agent model by learning {\it reward} it experiences.
%
Moreover, the IDEAL attention selection strategy has the flexibility to be
readily integrated with different deep learning features and
detectors therefore can benefit directly from models rapidly developed elsewhere. 
{\bf (2)} We introduce a simple yet powerful deep re-id model based on 
the Inception-V3 architecture \cite{szegedy2016rethinking}.
This model is learned directly by the identity classification loss
rather than the more common pairwise based verification \cite{Li_DeepReID_2014b,Ahmed2015CVPR} 
or triplet loss function \cite{ding2015deep}.
This loss selection not only significantly simplifies training data batch construction 
(e.g. random sampling with no notorious tricks required 
\cite{krizhevsky2012imagenet}), 
but also makes our model more scalable in practice given a large size
training population or imbalanced training data from different camera views.
%
We conducted extensive experiments 
on two large auto-detected datasets 
CUHK03~\cite{Li_DeepReID_2014b}
and Market-1501~\cite{zheng2015scalable}
to demonstrate the advantages of the proposed IDEAL model
over a wide range (24) of contemporary and state-of-the-art person
re-id methods. 
\vspace{0.1cm}
\noindent {\bf Related Work }
Most existing re-id methods~\cite{PCCA_CVPR12,KISSME_CVPR12,PRD_PAMI13,CVPR13LFDA,Zhao_MidLevel_2014a,
	xiong2014person,wang2014person,wang2016pami,Li_DeepReID_2014b,ding2015deep,
	Anton_2015_CoRR,liao2015person,Liao_2015_ICCV} focus on
supervised learning of person identity-discriminative information.  
Representative learning algorithms include 
ranking by pairwise or list-wise constraints
\cite{Anton_2015_CoRR,chen2015relevance,wang2016pami,loy2013person},
discriminative subspace/distance metric learning~\cite{KISSME_CVPR12,
	PRD_PAMI13,CVPR13LFDA,xiong2014person,liao2015person,Liao_2015_ICCV,zhang2016learning},
and deep learning~\cite{ShiZLLYL15,ding2015deep,
	Li_DeepReID_2014b,Ahmed2015CVPR,xiao2016learning,ding2015deep,wangjoint}.
They typically require a large quantity of person bounding boxes and
inter-camera pairwise identity labels,
which is prohibitively expensive to collect manually. 
%
\noindent {\bf\em Automatic Detection in Re-ID: }
Recent works~\cite{Li_DeepReID_2014b,zheng2015scalable,zheng2016PRW,zheng2016PRW}
have started to use automatic person detection for re-id benchmark
training and test. 
Auto-detected person bounding boxes contain more noisy background
and occlusions with misaligned person cropping (Fig. \ref{Fig:1}),
impeding discriminative re-id model learning.
A joint learning of person detection and re-id was also investigated
\cite{xiao2016end}.
However, the problem of {\it post-detection} attention selection for
re-id studied in this work has not been addressed in the
literature. Attention selection can benefit independently from
detectors rapidly developed by the wider community. 
%
%
%
\noindent {\bf\em Saliency and Attention Selection in Re-ID: }
Most related re-id techniques are
localised patch matching~\cite{shen2015person,zheng2015partial,Gated_SCNN} and
saliency detection~\cite{zhao2013person,hanxiao2014GTS,SalienceReId_CVPR13,liu2016end}.
They are inherently unsuitable by design to 
cope with poorly detected person images, due to their
stringent requirement of tight bounding boxes around the whole person. 
In contrast, the proposed IDEAL model is designed precisely to
overcome inaccurate bounding boxes therefore can potentially benefit
all these existing methods.
%
%
\noindent {\bf\em Reinforcement Learning in Computer Vision: }
Reinforcement Learning (RL) \cite{mnih2015human} is a problem
faced by an agent that learns its optimal behaviour by trial-and-error interactions with a dynamic environment \cite{kaelbling1996reinforcement}.
The promise of RL is offering a way of guiding the agent learning by reward and punishment
without the need for specifying how the target tasks to be realised.
Recently, RL has been successfully applied to a few vision tasks such as object localisation~\cite{caicedo2015active,mathe2016reinforcement,bellver2016hierarchical,jie2016tree}, image captioning
\cite{rennie2016self,liu2016optimization}, active object recognition \cite{malmir2017deep}.
To our best knowledge, this is the first attempt to 
exploit reinforcement learning 
for person re-id.
Compared to the most related fully supervised object localisation by RL
\cite{caicedo2015active,mathe2016reinforcement,bellver2016hierarchical,jie2016tree}, 
the proposed IDEAL model requires no accurate object bounding box annotations, 
therefore more scalable to large size data in practice.

\vspace{-0.5cm}
\section{Re-ID Attention Selection by Reinforcement Learning}
\label{sec:method}
\vspace{-0.2cm}
The {I}dentity {D}iscriminativ{E} {A}ttention reinforcement {L}earning
({IDEAL}) model has two sub-networks:
{\bf (I)} A multi-class {\em discrimination network} $\mathcal{D}$ by
deep learning from a training set of 
auto-detected person bounding boxes (Fig. \ref{Fig:architecture}(a)). This part is flexible with many options from existing
deep re-id networks and beyond
\cite{ding2015deep,wangjoint,xiao2016learning,wang2016highly}. 
%
%
{\bf (II)} A re-identification {\em attention network} $\mathcal{A}$ by reinforcement learning
recursively a salient sub-region with its deep feature
representation from $\mathcal{D}$ that can maximise identity-matching given
re-id label constraints (Fig. \ref{Fig:architecture}(b)). 
Next, we formulate the attention network by reinforcement learning and
how this attention network cooperates with the multi-class discrimination network.

\vspace{-0.3cm}
\subsection{Re-ID Attention Selection Formulation}
\label{sec:DQN}

We formulate the re-id attention selection
as a reinforcement learning problem \cite{kaelbling1996reinforcement}.
This allows to correlate directly the re-id attention selection process
with the learning objective of an ``agent'' by recursively {\em
  rewarding} or punishing the learning process.
In essence, the aim of model learning is to 
achieve an optimal identity discriminative attending action policy $a =
\pi(\mathbf{s})$ of an agent, 
i.e. a mapping function, that projects a state observation $\mathbf{s}$ (model input)
to an action prediction $a$.
In this work, we exploit the Q-learning technique
for learning the proposed IDEAL agent,
due to its sample efficiency advantage for a small set of actions \cite{watkins1989learning,gu2016q}.
Formally, we aim to learn an optimal state-value function
which measures the maximum sum of the current reward ($R_t$)
and all the future rewards ($R_{t+1}, R_{t+2}, \cdots$) 
discounted by a factor 
$\gamma$ at each time step $t$:
\begin{equation} \label{eq:Q}
Q^*(\mathbf{s}, a) = \max_\pi \mathbb{E}\big[R_t + \gamma R_{t+1} + \gamma^2R_{t+2} + \cdots \;\; | \;\; \mathbf{s}_t = \mathbf{s}, 
a_t = a, \pi \big]
\end{equation}
Once $Q^*(\mathbf{s}, a)$
is learned, the optimal policy 
$\pi^*(\mathbf{s})$ can be directly 
inferred by selecting the action
with the maximum $Q^*(\mathbf{s}, a)$ value in model deployment.
%
%
%
%
%
%
%
More specifically, the reinforcement learning agent interacts with each data sample
in a sequential episode,
which can be considered as a Markov decision process (MDP)~\cite{Puterman1994MDP}.
For our purpose, we need to design a specific MDP for re-id discriminative attention selection, as described below.

\vspace{-0.3cm}
\subsection{Markov Decision Process for Re-ID Attention Selection}
\label{sec:ideal}
We design a MDP for re-id attention selection
in auto-detected bounding boxes. 
In particular, we consider each input person bounding box image
as a dynamic environment.
An IDEAL agent interacts with this dynamic environment
to locate the optimal re-id attention window.
To guide this discriminative learning process, 
we further consider a reward that can encourage those attending actions 
to improve re-id performance and
maximise the cumulative future reward in Eqn.~\eqref{eq:Q}.
As such, we define actions, states, and rewards
as follows.




\begin{figure}
\centering
\includegraphics[width=1\linewidth]{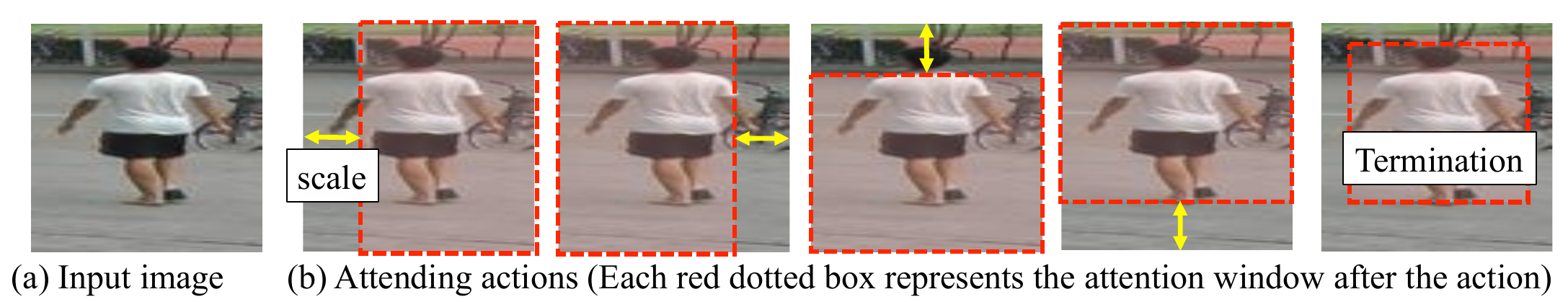}
\vskip -0.4cm
\caption{\footnotesize
	Identity discriminative attending actions are
  given by an attending scale variable on four directions (left/right/top/bottom).
	Termination action means the stop of a recursive attending process.
}
\label{Fig:actions}
\end{figure}

\vspace{0.1cm}
\noindent \textbf{Actions}: 
An action set $\mathbf{A}$ is defined to facilitate the IDEAL agent to
determine the location and size of an ``attention window'' (Fig.~\ref{Fig:actions}).
%
Specifically, an attending action $a$ is defined by
the location shift direction ($a_d \in \{\text{left, right, top, bottom}\}$) 
and shift scale ($a_e \in \mathbf{E}$).
We also introduce a termination action 
as a search process stopping signal.
$\mathbf{A}$ consists of a total of ($4\times |\mathbf{E}|+1$) actions.
Formally, 
let the upper-left and bottom-right corner coordinates of the current attention window
and an updated attention window be $\left[x_1, y_1, x_2, y_2\right]$ and
$\left[ x_1', y_1', x_2', y_2' \right]$ respectively,
the action set $\mathbf{A}$ can then be defined as:
\begin{align}
&\mathbf{A} = \{ x_1'= x_1 + \alpha \Delta x, \;\; x_2'=x_2 - \alpha \Delta x, \;\; y_1'=y_1 + \alpha \Delta y, \;\; y_2'= y_2 - \alpha \Delta y,  \;\; \text{T}\} , \\
&\text{where} \;\;\; \alpha \in \mathbf{E}, 
\;\;\;
\Delta x = x_2 - x_1, \;\;\;
\Delta y = y_2 - y_1, \;\; 
\text{T} = \text{termination} . \nonumber
\end{align}
Computationally, each action except termination in $\mathbf{A}$ modifies the environment by cutting off a horizontal or vertical stripe.
We set $\mathbf{E} = \{5\%, 10\%, 20\%\}$ by cross-validation in 
our experiments,
resulting in total 13 actions.
Such a small attention action space with multi-scale changes has three merits:
(1) Only a small number of simple actions are contained, which allows
more efficient and stable agent training;
(2) Fine-grained actions with small attention changes allow 
the IDEAL agent sufficient freedoms
to utilise small localised regions in
auto-detected bounding boxes for subtle identity matching.
This enables more effective elimination of undesired background
clutter whilst retaining identity discriminative information;
(3) The termination action enables the agent to be aware of
the satisfactory condition met for attention selection and stops
further actions when optimised.

\vspace{0.1cm}
\noindent \textbf{States}: 
The state $\mathbf{s}_t$ of our MDP at time $t$ is defined as
the concatenation of the feature vector $\mathbf{x}_t \in \mathbb{R}^d$ 
(with $d$ re-id feature dimension)
of current attending window
and an action history vector $\mathbf{h}_t \in \mathbb{R}^{|\mathbf{E}| \times n_\text{step}}$ 
(with $n_\text{step}$ a pre-defined maximal action number per bounding box),
i.e. $\mathbf{s}_t = \left[ \mathbf{x}_t, \mathbf{h}_t \right]$.
Specifically, at each time step, we extract the feature vector $\mathbf{x}_t$ 
of current attention window by the trained {re-id network}
$\mathcal{D}$.
The action history vector $\mathbf{h}_t$ is a binary
vector for keeping a track of 
all past actions,
represented by a $|\bm{A}|$-dimensional 
(13 actions)
one-hot vector where the corresponding action bit
is encoded as one, all others as zeros.
 
\vspace{0.1cm}
\noindent \textbf{Rewards}:
The reward function $R$ (Eqn. \eqref{eq:Q}) defines the agent task objective.
In our context, we therefore correlate directly the reward function 
of the IDEAL agent's attention behaviour with the re-id matching criterion.
Formally, at time step $t$, 
suppose the IDEAL agent observes a
person image $\mathbf{I}_t$ and then takes
an action $a_t = a \in \mathbf{A}$ to
attend the image region $\mathbf{I}_t^a$. 
Given this attention shift from
$\mathbf{I}_t$ to $\mathbf{I}_t^a$,
its state $\mathbf{s}_t$ changes to 
$\mathbf{s}_{t+1}$.
We need to assess such a state change and signify 
the agent if this action is encouraged or 
discouraged by an award or a punishment.
To this end, we propose three reward function designs, 
inspired by pairwise constraint learning principles
established in generic information search and person re-id.

\vspace{0.1cm}
\noindent {\em Notations }
From the labelled training data, we sample two other {\it reference} images
w.r.t. $\mathbf{I}_t$:
(1) A {\it cross-view positive} sample $\mathbf{I}_t^+$
sharing the same identity as $\mathbf{I}_t$ but not the camera view; 
(2) A {\it same-view negative} sample $\mathbf{I}_t^-$
sharing the camera view as $\mathbf{I}_t$ but not the identity.
We compute the features of all these images by $\mathcal{D}$,
denoted respectively as $\mathbf{x}_t, \mathbf{x}_t^a, \mathbf{x}_t^+$,
and $\mathbf{x}_t^-$.

\vspace{0.1cm}
\noindent {\bf\em (I) Reward by Relative Comparison }
Our first reward function $R_{t}$ is based on relative comparison,
in spirit of the triplet loss for learning to rank \cite{liu2009learning}.
It is formulated as:
\begin{equation} 
\label{eq:_Hierachical_2}
R_t=R_{rc}(\mathbf{s}_t,a)
= \Big(f_\text{match}(\mathbf{x}_t^a, \mathbf{x}_t^-) - 
f_\text{match}(\mathbf{x}_t^a, \mathbf{x}_t^+)\Big) - 
\Big(f_\text{match}(\mathbf{x}_t, \mathbf{x}_t^-) - 
f_\text{match}(\mathbf{x}_t, \mathbf{x}_t^+)\Big)
\end{equation}
where $f_\text{match}$ defines the re-id matching function.
We use the Euclidean distance metric given the Inception-V3 deep features.
Intuitively, this reward function commits (i) a positive reward if the
attended region becomes more-matched to the {\it cross-view positive}
sample whilst less-matched to the {\it same-view negative} sample, or (ii) a negative reward otherwise. 
When $a$ is the termination action, i.e. $\mathbf{x}^a_t = \mathbf{x}_t$, 
the reward value $R_{rc}$ is set to zero. 
In this way, the IDEAL agent is supervised to attend the regions
subject to optimising jointly two tasks: 
(1) being more discriminative and/or more salient for
the target identity in an inter-view sense (cross-view re-id),
whilst (2) pulling the target identity further away 
from other identities in an intra-view sense
(discarding likely shared view-specific background clutter and occlusion
therefore focusing more on genuine person appearance).
Importantly, this multi-task objective design favourably allows appearance saliency learning to intelligently select the most informative parts of
certain appearance styles for enabling holistic clothing patten detection and ultimately
more discriminative re-id matching (e.g. Fig. \ref{Fig:1}(b) and 
Fig. \ref{Visuliaztion_DQN_V2_training}(b)).

\vspace{0.1cm}
\noindent {\bf\em (II) Reward by Absolute Comparison }
Our second reward function considers only the compatibility of a true matching pair,
in the spirit of positive verification constraint learning \cite{chopra2005learning}.
Formally, this reward is defined as:
\begin{equation} 
\label{eq:_Hierachical_3}
R_t=R_{ac}(\mathbf{s}_t,a)
=  \Big( f_\text{match}(\mathbf{x}_t, \mathbf{x}_t^+) \Big) - 
\Big( f_\text{match}(\mathbf{x}_t^a, \mathbf{x}_t^+)\Big)
\end{equation}
The intuition is that, the cross-view matching score of two same-identity 
images depends on how well irrelevant background clutter/occlusion is
removed by the current action.  
That is, a good attending action will increase a cross-view matching
score, and vice verse.

\vspace{0.1cm}
\noindent {\bf\em (III) Reward by Ranking }
Our third reward function concerns the true match ranking change brought by the agent action, 
therefore simulating directly the re-id deployment rational \cite{REIDchallenge}.
Specifically, 
we design a binary reward function according to
whether the rank of true match $\mathbf{x}_{t}^{+}$ 
is improved when $\mathbf{x}_{t}$ and $\mathbf{x}_{t}^{a}$ are used as the probe separately, as:  
\begin{equation} 
R_t=R_{r}(\mathbf{s}_t,a)=\left\{ 
\begin{array}{lr}  
+1,  \ \ \text{if} \;\; 
\text{Rank}(\mathbf{x}_{t}^{+}|\mathbf{x}_{t}) >
\text{Rank}(\mathbf{x}_{t}^{+}|\mathbf{x}_{t}^a)\\
-1,   \  \ \text{otherwise}\\  
\end{array}  
\right.  
\label{eqn:ranking}
\end{equation}
where $\text{Rank}(\mathbf{x}_{t}^{+}|\mathbf{x}_{t})$ 
($\text{Rank}(\mathbf{x}_{t}^{+}|\mathbf{x}_{t}^a)$) represents the rank
of $\mathbf{x}_{t}^{+}$ in a gallery 
against the probe $\mathbf{x}_{t}$ ($\mathbf{x}_{t}^a$).
Therefore, Eqn.~\eqref{eqn:ranking} gives support to those actions of
leading to a higher rank for the true match, which is precisely the re-id objective.
In our implementation, the gallery was constructed by randomly sampling
$n_g$ (e.g. 600) cross-view training samples.
We evaluate and discuss the above three reward function choices in 
the experiments (Sec.~\ref{sec:exp}).

%
%
%
%


\vspace{-0.3cm}
\subsection{Model Implementation, Training, and Deployment}

\noindent {\bf Implementation and Training } 
For the multi-class discrimination network $\mathcal{D}$ in the IDEAL
model, we deploy the Inception-V3
network~\cite{szegedy2016rethinking} (Fig. \ref{Fig:architecture}(a)), 
a generic image classification CNN model \cite{szegedy2016rethinking}.
It is trained from scratch by a softmax classification loss using person identity labels
of the training data.
%
For the re-id attention network $\mathcal{A}$ in the IDEAL model, we design a neural network of
3 fully-connected layers (each with 1024 neurons) and a prediction layer
(Fig. \ref{Fig:architecture}(b)). This implements the state-value function Eqn.~\eqref{eq:Q}.
For optimising the sequential actions for re-id attention selection,
we utilise the $\epsilon$-greedy learning algorithm \cite{mnih2015human} 
during model training:   
The agent takes (1) a random action
from the action set $\mathbf{A}$ with the probability $\epsilon$,
and (2) the best action predicted by
the agent with the probability $1-\epsilon$.
We begin with $\epsilon=1$ 
and gradually decrease it by $0.15$ every 1 training epoch until reaching $0.1$. The purpose is to balance model exploration and exploitation in the training stage
so that local minimum can be avoided.
To further reduce the correlations between sequential observations,
we employ the experience replay strategy \cite{mnih2015human}.
In particular,
a fixed-sized memory pool $\mathbf{M}$ is created to store the agent's
$N$ past training sample (experiences) $e_t = (\mathbf{s}_t, a_t, R_t,
\mathbf{s}_{t+1})$ at each time step $t$,  
i.e. $\{ e_{t-N+1}, \cdots, e_t \}$.
At  iteration $i$, 
a mini-batch of training samples is selected randomly from
$\mathbf{M}$ to update the agent parameters $\mathbf{\theta}$ by the
loss function: 
\begin{equation}
L_i(\mathbf{\theta}_i) = \mathbb{E}_{(\mathbf{s}_t, a_t, R_{t}, \mathbf{s}_{t+1})\sim \text{Uniform}(\mathbf{M})} 
\Big( R_{t} + \gamma \max_{a_{t+1}}Q(\mathbf{s}_{t+1}, a_{t+1}; \tilde{\mathbf{\theta}}_i) - Q(\mathbf{s}_t, a_t; \mathbf{\theta}_i) \Big)^2,
\end{equation}
where $\tilde{\mathbf{\theta}}_i$ are the parameters of an intermediate model 
for predicting training-time target values, which are updated as 
$\mathbf{\theta}_i$ at every $\varsigma$ iterations, but frozen at other times.

\vspace{0.1cm}
\noindent {\bf Deployment } 
During model deployment,
we apply the learned attention network $\mathcal{A}$ 
to all test probe and gallery bounding boxes for extracting their
attention window images.
The deep features of these attention window images are used
for person re-id matching by extracting the 2,048-D output from the
last fully-connected layer of the discrimination network $\mathcal{D}$. 
We employ the L2 distance as the re-id matching metric.

\vspace{-0.3cm}
\section{Experiments}
\label{sec:exp}

\noindent \textbf{Datasets }
For evaluation, we used two large benchmarking re-id datasets generated
by automatic person detection:
\label{key}CUHK03 \cite{Li_DeepReID_2014b}, and Market-1501 \cite{zheng2015scalable}
(details in Table \ref{tab:datasets}).
CUHK03 also provides an extra version of bounding boxes by human labelling
therefore offers a like-to-like comparison between 
the IDEAL attention selection and human manually cropped images.
Example images are shown in (a),(b) and (c) of Fig.~\ref{Fig:1}. 

\noindent \textbf{Evaluation Protocol }
We adopted the standard CUHK03 1260/100 
\cite{Li_DeepReID_2014b} and
Market-1501 750/751 
\cite{zheng2015scalable} training/test person split.
We used the single-shot setting on CUHK03,
both single- and multi-query setting on Market-1501.
We utilised the cumulative matching characteristic (CMC) to measure re-id
accuracy. For Market-1501, we also used the recall
measure of multiple truth matches by mean Average Precision (mAP).

\noindent \textbf{Implementation Details }
We implemented the proposed IDEAL method in 
the TensorFlow framework \cite{abadi2016tensorflow}.
We trained an Inception-V3 \cite{szegedy2016rethinking} 
multi-class identity discrimination network $\mathcal{D}$ from scratch
for each re-id dataset at a learning rate of 0.0002 by using the Adam
optimiser \cite{kingma2014adam}. The final FC layer output feature
vector (2,048-D) together with the L2 distance metric is used as our
re-id matching model. 
All person bounding boxes were resized to $299 \times 299$ in pixel.
We trained the $\mathcal{D}$ by 100,000 iterations. 
%
%
%
We optimised the IDEAL attention network $\mathcal{A}$ 
by the Stochastic Gradient Descent algorithm \cite{bottou2012stochastic}
with the learning rate set to 0.00025.
We used the relative comparison based reward function (Eqn.~\eqref{eq:_Hierachical_2}) by default.
The experience replay memory ($\mathbf{M}$) size for reinforcement learning was 100,000.
We fixed the discount factor $\gamma$ to 0.8 (Eqn. \eqref{eq:Q}). 
We allowed a maximum of $n_\text{step}=5$ action rounds 
for each episode in training $\mathcal{A}$. 
The intermediate regard prediction network was updated every $\varsigma=100$ iterations. 
We trained the $\mathcal{A}$ by 10 epochs. 


%

\begin{table}[h]
	\centering
	\scriptsize
	\renewcommand{\arraystretch}{1}
	\setlength{\tabcolsep}{0.05cm}
	\caption{ \footnotesize
		Comparing re-id performance. 
		$1^\text{st}/2^\text{nd}$ best results 
		are shown in red/blue. AD: Automatically Detected.
	}
	\vskip -0.3cm
	\label{tab:CMC_state_of_art}
	\scalebox{1}{
		\begin{tabular}{|c|cccc|cc|cc||c|cccc|cc|cc|}
			\whline
			Dataset  & \multicolumn{4}{c|}{CUHK03(AD) \cite{Li_DeepReID_2014b}} & 
			\multicolumn{4}{c||}{Market-1501(AD) \cite{zheng2015scalable}}
			&   & \multicolumn{4}{c|}{CUHK03(AD) \cite{Li_DeepReID_2014b}} & 
			\multicolumn{4}{c|}{Market-1501(AD) \cite{zheng2015scalable}}
			\\ \hline 
			\multirow{2}{*}{Metric (\%)} & 
			\multirow{2}{*}{R1} & \multirow{2}{*}{R5} & \multirow{2}{*}{R10} & \multirow{2}{*}{R20} &
			\multicolumn{2}{c|}{Single Query} & \multicolumn{2}{c||}{Multi-Query} & 
			& \multirow{2}{*}{R1} & \multirow{2}{*}{R5} & \multirow{2}{*}{R10} & \multirow{2}{*}{R20} &
			\multicolumn{2}{c|}{Single Query} & \multicolumn{2}{c|}{Multi-Query} \\
			& & & & & R1 & mAP & R1 & mAP & & & & &
			& R1& mAP & R1 & mAP
			\\ 
			\hline \hline
			ITML\cite{ITM_ICML07} & 5.1 & 17.7 & 28.3 & - & -& -&- &- &
			TMA\cite{martinel2016temporal} & -& -& -& -& 47.9 & 22.3 & - & - \\
			LMNN\cite{LMNN_JMLR09} & 6.3 & 18.7 & 29.0 & - &  -&- &- &- &
			HL\cite{ustinova2016learning} &- & -& -& - & 59.5 & - & - & - \\ 
			KISSME\cite{KISSME_CVPR12} & 11.7 & 33.3 & 48.0 & - & 40.5 & 19.0 & - & -  &
			HER\cite{wang2016highly} & 60.8 & 87.0 & \color{red} \bf 95.2 & \color{red} \bf 97.7 & - & - & - & - \\
			\cline{10-18}
			MFA\cite{xiong2014person} &- &- & -& -& 45.7 & 18.2 & - & - &
			FPNN\cite{Li_DeepReID_2014b} & 19.9 & - & - & -  &- &- & -&- \\
			kLFDA\cite{xiong2014person} &- & -& -& -& 51.4 & 24.4 & 52.7 & 27.4 &
			DCNN+\cite{Ahmed2015CVPR}& 44.9 & 76.0 & 83.5  & 93.2  & -&- &- &-
			\\
			BoW\cite{zheng2015scalable} & 23.0 & 42.4 & 52.4 & 64.2 & 34.4 & 14.1 & 42.6 & 19.5 &
			EDM\cite{shi2016embedding} & 52.0 & - & -  &-  &- & -&- &- 
			\\
			XQDA\cite{liao2015person}& 46.3 & 78.9 & 83.5  & 93.2 &43.8 & 22.2 & 54.1 & 28.4 &
			SICI\cite{wangjoint}  & 52.1 & 84.9 & 92.4  & - &- & -& -&-
			\\ 
			MLAPG\cite{Liao_2015_ICCV}& 51.2 & 83.6 & 92.1  & \color{blue} \bf 96.9  & -&- &- &- &
			SSDAL\cite{su2016deep}&- &- &- &- & 39.4 & 19.6 & 49.0 & 25.8 
			\\ 
			L$_1$-Lap~\cite{ElyorECCV16} & 30.4 & - & - & -  & -&- &- &- &
			S-LSTM~\cite{S_LSTM} & 57.3 & 80.1 & 88.3 & - & - & - & 61.6 & 35.3 \\
			\cline{10-18}
			NFST\cite{zhang2016learning}& 53.7 & 83.1 & 93.0  & 94.8 & 55.4 & 29.9 & 68.0 & 41.9 &
			eSDC\cite{SalienceReId_CVPR13}
			& 7.7 & 21.9 & 35.0 & 50.0 & 33.5 & 13.5 & - & -
			\\ 
			LSSCDL\cite{LSSCDL}& 51.2 & 80.8 & 89.6 & - & -&- &- &- &
			CAN\cite{liu2016end}
			& 63.1 & 82.9 & 88.2 & 93.3 & 48.2 & 24.4 & - & -
			\\
			 
			SCSP\cite{chen2016similarity} & -& -&- &- & 51.9 & 26.3 & - & - &
			GS-CNN\cite{Gated_SCNN} & \bf \color{blue} 68.1 & \bf \color{blue} 88.1 & \color{blue} \bf{94.6}  & - & \bf \color{blue} 65.8 & \bf \color{blue} 39.5 & \bf \color{blue} 76.0 & \bf \color{blue} 48.4  \\
			\hline
			& & & & & & & & &
			{\bf IDEAL} 
			& \color{red} \bf{71.0} & \color{red} \bf{89.8} & 93.0 & {95.9}  
			& \color{red} \bf{86.7} & \color{red} \bf{67.5} & \color{red} \bf{91.3} & \color{red} \bf{76.2} \\ 
			\whline
	\end{tabular}
}
\end{table}
\begin{figure}[h]
	\centering
	\includegraphics[width=1\linewidth]{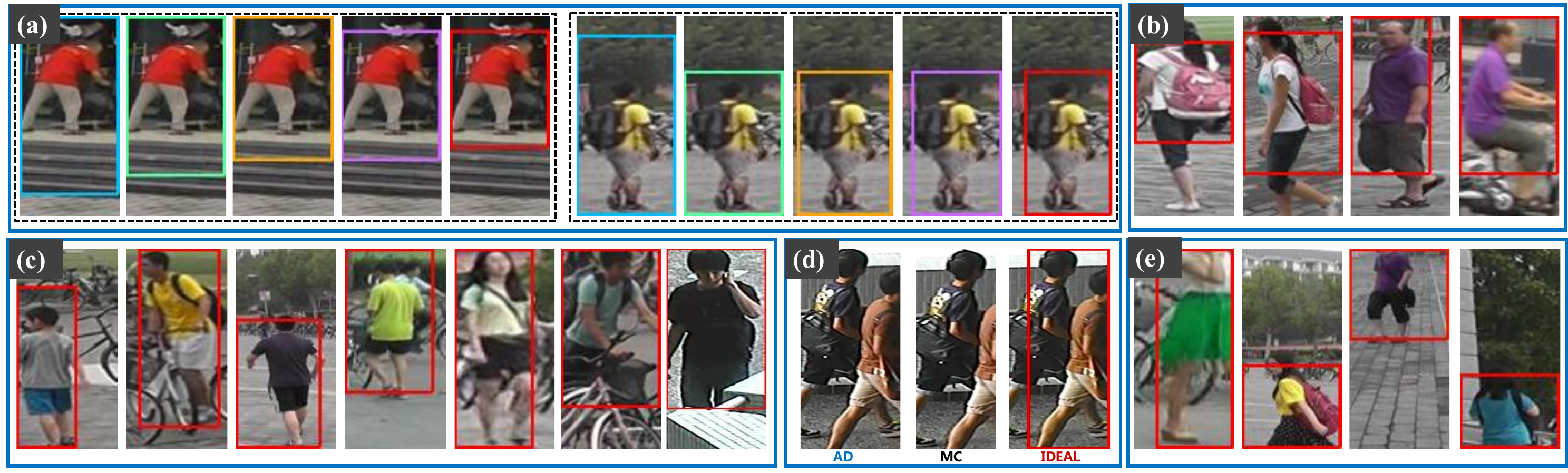} 
	\vskip -0.4cm
	\caption{\footnotesize
		Qualitative evaluations of the IDEAL model:
		{\bf (a)} Two examples of action sequence for attention selection given by
		action1 (Blue), action2 (Green), action3 (Yellow),
		action4 (Purple), action5 (Red);
		{\bf (b)} Two examples of cross-view IDEAL selection for re-id;
		{\bf (c)} Seven examples of IDEAL selection given by 5, 3, 5, 5, 4, 2, and 2 
		action steps respectively;
		{\bf (d)} A failure case when the original auto-detected (AD) bounding box
         contains two people, manually cropped (MC) gives a more
         accurate box whilst IDEAL attention selection fails to reduce
         the distraction; 
		{\bf (e)} Four examples of IDEAL selection on the Market-1501 ``distractors''
		 with significantly poorer auto-detected bounding
                 boxes when IDEAL shows greater effects.
	}
	\label{Visuliaztion_DQN_V2_training}
\end{figure}

\noindent{\bf Comparisons to the State-of-the-Arts}
%
We compared the IDEAL model against 24 different contemporary and the
state-of-the-art re-id methods (Table \ref{tab:CMC_state_of_art}). It
is evident that IDEAL achieves the best
re-id performance, 
outperforming the strongest competitor GS-CNN~\cite{Gated_SCNN} by 
$2.9\%$ (71.0-68.1) and 
$20.9\%$ (86.7-65.8) in Rank-1 on CUHK03 and Market-1501 respectively.
This demonstrates a clear positive effect of IDEAL's attention selection on person
re-id performance by filtering out bounding
box misalignment and random background clutter in auto-detected person images. To give more insight and
visualise both the effect of IDEAL and also failure cases, qualitative examples are
shown in Fig.~\ref{Visuliaztion_DQN_V2_training}.

\begin{table}[h!]
	\centering
	\footnotesize
	\renewcommand{\arraystretch}{1}
	\setlength{\tabcolsep}{0.1cm}
	\caption{
		Comparing attention selection methods.
		SQ: Single Query; MQ: Multi-Query.
	}
	\vskip -0.3cm
	\label{tab:Attention Strategy}
	\scalebox{1}{
		\begin{tabular}{|c|cccc|cc|cc|}
			\whline
			Dataset  & \multicolumn{4}{c|}{CUHK03 \cite{Li_DeepReID_2014b}} & 
			\multicolumn{4}{c|}{Market-1501 \cite{zheng2015scalable}}\\ \hline 
			Metric (\%) & R1 & R5 & R10 & R20 & R1(SQ) & mAP(SQ) & R1(MQ)& mAP(MQ)\\ \hline \hline
			eSDC \cite{SalienceReId_CVPR13}
			& 7.7 & 21.9 & 35.0 & 50.0 & 33.5 & 13.5 & - & - \\
			CAN \cite{liu2016end}
			& 63.1 & 82.9 & 88.2 & 93.3 & 48.2 & 24.4 & - & - \\
			GS-CNN~\cite{Gated_SCNN} & 68.1 & 88.1 & \bf{94.6}  & - & 65.8 & 39.5 & 76.0 & 48.4 \\
			\hline
			No Attention & 64.9 & 84.5& 92.6 & 95.7 &84.5 & 64.8 &89.4 &  72.5
			\\ \hline
			Random Attention & 54.1 &79.2  &85.9  &90.4  &80.3 &54.6 &85.1&66.7 
			\\ \hline
			Centre Attention (95\%) & 66.1 & 86.7 &91.1  & 94.9 &  84.1& 64.2&88.6 &69.4   \\
			Centre Attention (90\%) & 64.1&85.3&90.3&93.5&82.7&60.3&87.5&65.3\\
			Centre Attention (80\%) &51.9&76.0&83.0&89.0&74.7&48.5&83.4&57.6\\
			Centre Attention (70\%) &35.2&62.3&73.2&81.7&63.8&39.0&72.3&43.5\\
			Centre Attention (50\%) &16.7&38.8&49.5&62.5 &39.9&18.5&46.3&23.9 \\
			\hline
			\bf{IDEAL(Ranking)} &70.3  & 89.1 & 92.7 &95.4  &  86.2&66.3 & 90.8 & 74.3  \\
			\bf{IDEAL(Absolute Comparison)}&  69.1& 88.4  & 92.1 & 95.0 & 85.3 & 65.5 & 87.5&72.3  \\		
			\bf{IDEAL(Relative Comparison)} 
			&\bf{71.0}&\bf{89.8} &{93.0} &\bf{95.9}  &\bf {86.7}& \bf{67.5}  & \bf{91.3}&\bf{76.2}
			\\ \whline
	\end{tabular}}
\end{table}

\noindent{\bf Evaluations on Attention Selection }
%
We further compared in more details the IDEAL model against
three state-of-the-art saliency/attention based re-id models 
(eSDC \cite{SalienceReId_CVPR13},
CAN \cite{liu2016end}, GS-CNN \cite{Gated_SCNN}), and
two baseline attention methods (Random, Centre) using 
the Inception-V3 re-id model (Table \ref{tab:Attention Strategy}).
For {\em Random Attention},
we attended randomly person bounding boxes 
by a ratio (\%) randomly selected from $\{95,90,80,70,50\}$.
We repeated 10 times and reported the mean results.
For {\em Centre Attention},
we attended all person bounding boxes at centre by 
one of the same 5 ratios above. 
%
It is evident that the IDEAL (Relative Comparison) model is the best. %
The inferior re-id performance of eSDC, CAN and GS-CNN is
due to their strong assumption
on accurate bounding boxes. 
Both Random and Centre Attention methods do not work either with even
poorer re-id accuracy than that with ``No Attention'' selection. 
This demonstrates that optimal attention selection given by IDEAL is non-trivial.
Among the three attention reward functions, 
{\em Absolute Comparison} is the weakest,
likely due to
the lack of reference comparison against false matches,
i.e. no population-wise matching context in attention learning.
{\em Ranking} fares better, as it considers reference comparisons. 
The extra advantage of {\em Relative Comparison} is due to the
{\em same-view negative} comparison in Eqn.\eqref{eq:_Hierachical_2}. This provides a more reliable
background clutter detection since 
same-view images are more likely to share similar background
patterns. 

\noindent {\bf Auto-Detection+IDEAL vs. Manually Cropped }
Table \ref{tab:compare_labelling} shows that
auto-detection+IDEAL can perform similarly to that of {\em manually
  cropped} images in CUHK03 test\footnote{The Market-1501 dataset
  provides no manually cropped person bounding boxes.},
e.g. $71.0\%$ vs. $71.9\%$ for Rank-1 score.
This shows the potential of IDEAL in eliminating 
expensive manual labelling of bounding boxes and 
for scaling up re-id to large data deployment. 


\vspace{-0.3cm}
\begin{table}[h]
	\centering
	\footnotesize
	\caption{
		Auto-detection+IDEAL vs. manually cropped re-id on CUHK03.
	}
	\vskip -0.3cm
	\label{tab:compare_labelling}
	\scalebox{1}{
		\renewcommand{\arraystretch}{1}
		\setlength{\tabcolsep}{0.26cm}
		
		\begin{tabular}{|c||cccc|}
			\whline
			Metric (\%) & R1 & R5 & R10 & R20 \\ \hline 
			%
			Auto-Detected+{\bf IDEAL} &  {71.0} & {89.8} & 93.0 & {95.9} 
			\\ \cline{1-5}
			Manually Cropped &\bf 71.9&\bf{90.4} &\bf{94.5} & \bf{97.1}\\	
			\whline
	\end{tabular}}
\end{table}
\vspace{-0.3cm}

\vspace{0.1cm}
\noindent {\bf Effect of Action Design }
We examined three designs with distinct attention scales.
Table \ref{tab:action_scale} shows that 
the most fine-grained design $\{5\%, 10\%, 20\%\}$ is the best.
This suggests that the re-id by appearance is subtle 
and small regions make a difference in discriminative matching.

\vspace{-0.3cm}
\begin{table}[h]
	\centering
	\footnotesize
	\renewcommand{\arraystretch}{1}
	\setlength{\tabcolsep}{0.16cm}
	\caption{
		Attention action design evaluation. SQ: Single Query; MQ: Multi-Query.
	}
	\vskip -0.3cm
	\label{tab:action_scale}
	\scalebox{1}{	
		\begin{tabular}{|c|cccc|cc|cc|}
			\whline
			Dataset  & \multicolumn{4}{c|}{CUHK03 \cite{Li_DeepReID_2014b}} & 
			\multicolumn{4}{c|}{Market-1501 \cite{zheng2015scalable}}\\ \hline 
			Metric (\%) & R1 & R5 & R10 & R20 & R1(SQ) & mAP(SQ) & R1 (MQ)& mAP(MQ)\\ \hline \hline
			\{5\%, 10\%, 20\%\} &\bf 71.0&\bf 89.8 &\bf 93.0&\bf 95.9 
			&\bf 86.7&\bf 67.5 &\bf  91.3&\bf 76.2 \\	
			\hline
			\{10\%, 20\%, 30\%\} &68.3&88.1&91.8&95.0&86.2&66.8&90.5  &73.4   \\
			\hline
			\{10\%, 20\%, 50\%\} & 67.6&87.5 &91.4  &93.9&85.3&65.6&88.8&72.1     \\
			\whline
	\end{tabular}}
\end{table}
\vspace{-0.5cm}

\vspace{-0.4cm}
\section{Conclusion}
\label{sec:conclusion}

We presented an Identity DiscriminativE Attention reinforcement Learning (IDEAL) 
model for optimising re-id attention selection in auto-detected bounding boxes.
This improves notably person re-id accuracy in a fully automated
process required in practical deployments. 
The IDEAL model is formulated as a unified framework of 
discriminative identity learning by a deep multi-class discrimination network 
and attention reinforcement learning
by a deep Q-network.
This achieves jointly optimal identity sensitive attention selection
and re-id matching performance by a reward function subject to identity label pairwise constraints.
Extensive comparative evaluations on two auto-detected re-id
benchmarks show clearly the advantages and superiority of this IDEAL
model in coping with bounding box misalignment and background clutter
removal when compared to the state-of-the-art saliency/attention based
re-id models. Moreover, this IDEAL automatic attention selection
mechanism comes near to be equal to human manual labelling of person
bounding boxes on re-id accuracy, 
therefore showing a great potential for scaling up automatic re-id to
large data deployment.


\bibliography{reid_bmvc17}

\end{document}